%% file: main.tex
\documentclass[10pt,twocolumn,letterpaper]{article}

\usepackage{cvpr}              
\input{preamble}
\definecolor{cvprblue}{rgb}{0.21,0.49,0.74}
\usepackage[pagebackref,breaklinks,colorlinks,allcolors=cvprblue]{hyperref}

\def\paperID{10112} 
\def\confName{CVPR}
\def\confYear{2026}

\title{InterPhys: Physics-aware Human Motion Synthesis in a Dynamic Scene}

\author{
Chaoyue Xing \quad
Wei Mao \quad
Miaomiao Liu\\[0.5em]
Australian National University, Canberra, Australia\\
{\tt\small \{chaoyue.xing,   miaomiao.liu\}@anu.edu.au;\;wei.mao.research@gmail.com }
}

\begin{document}
\maketitle
\input{sec/01-abstract}    
\input{sec/02-introduction}
\input{sec/03-related-work}

\input{sec/04-method}

\input{sec/05-experiments}
\input{sec/06-conclusion}
{
    \small
    \bibliographystyle{ieeenat_fullname}
    \bibliography{egbib}
}


\end{document}

%% file: sec/01-abstract.tex
\begin{abstract}
This paper tackles the problem of physics-aware human motion synthesis in a dynamic scene. Unlike existing works which mainly tend to generate physically unrealistic motions due to limited contact modeling, typically restricted to hands, in this paper, we introduce a physics-aware human motion generation framework that explicitly models the full spectrum of human-related forces, including human-object, human-scene, and internal body dynamics.~Our method imposes soft physical constraints to maintain force and torque balance, ensuring physically grounded motion synthesis. We further propose a novel continuous distance-based force model that generalizes contact modeling to arbitrary surfaces, capturing interactions not only with static environments but also with dynamic, moving objects. Extensive experiments show that our approach significantly improves physical plausibility and generalizes well to complex scenes, setting a new benchmark for physically consistent human motion generation.  
\end{abstract}

%% file: sec/02-introduction.tex
\begin{figure*}[t]
\centering
   \begin{tabular}{cc}
    \hspace{-2mm}  \includegraphics[width=0.48\linewidth]{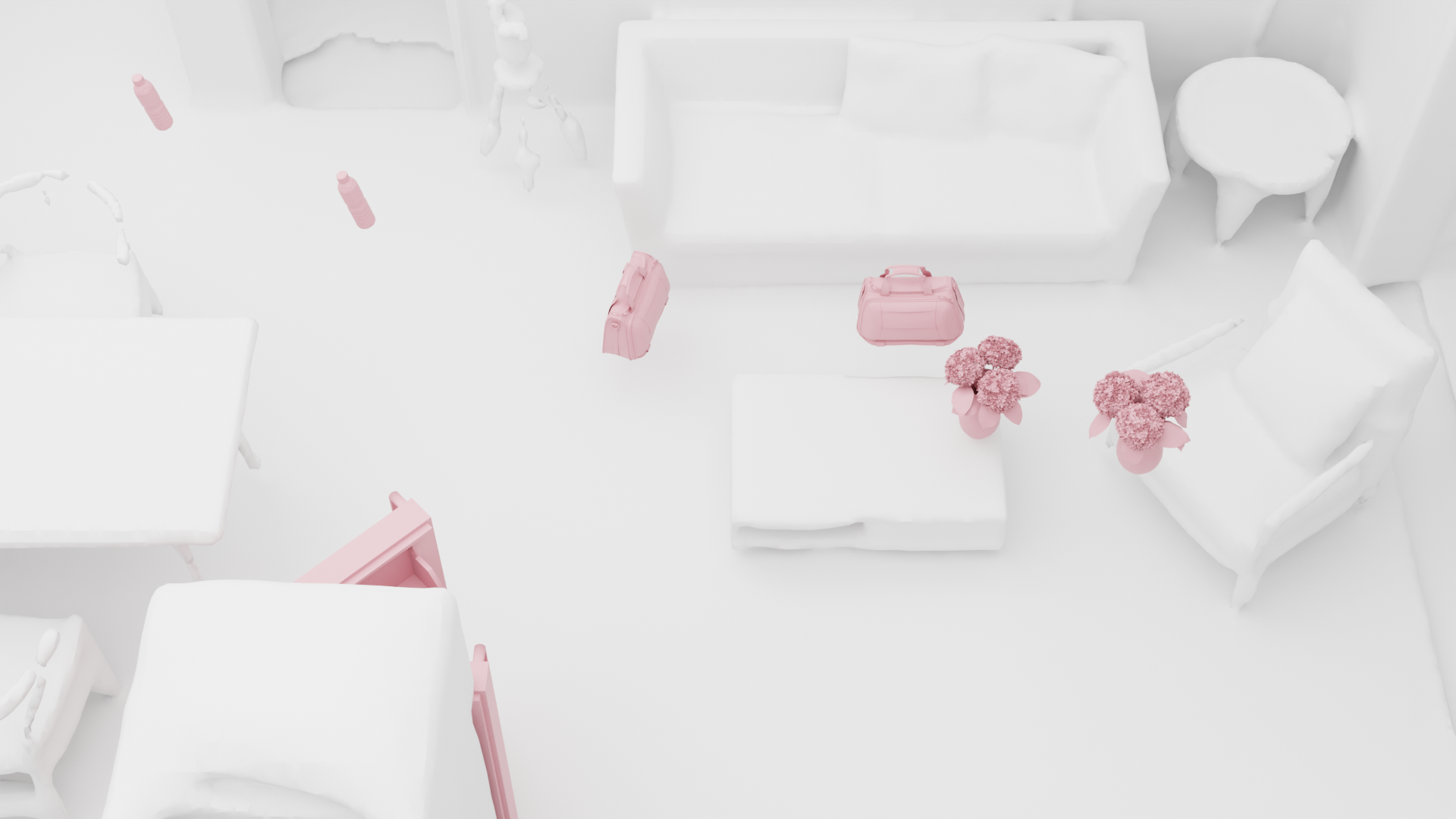}  &  \hspace{-2mm}\includegraphics[width=0.48\linewidth]{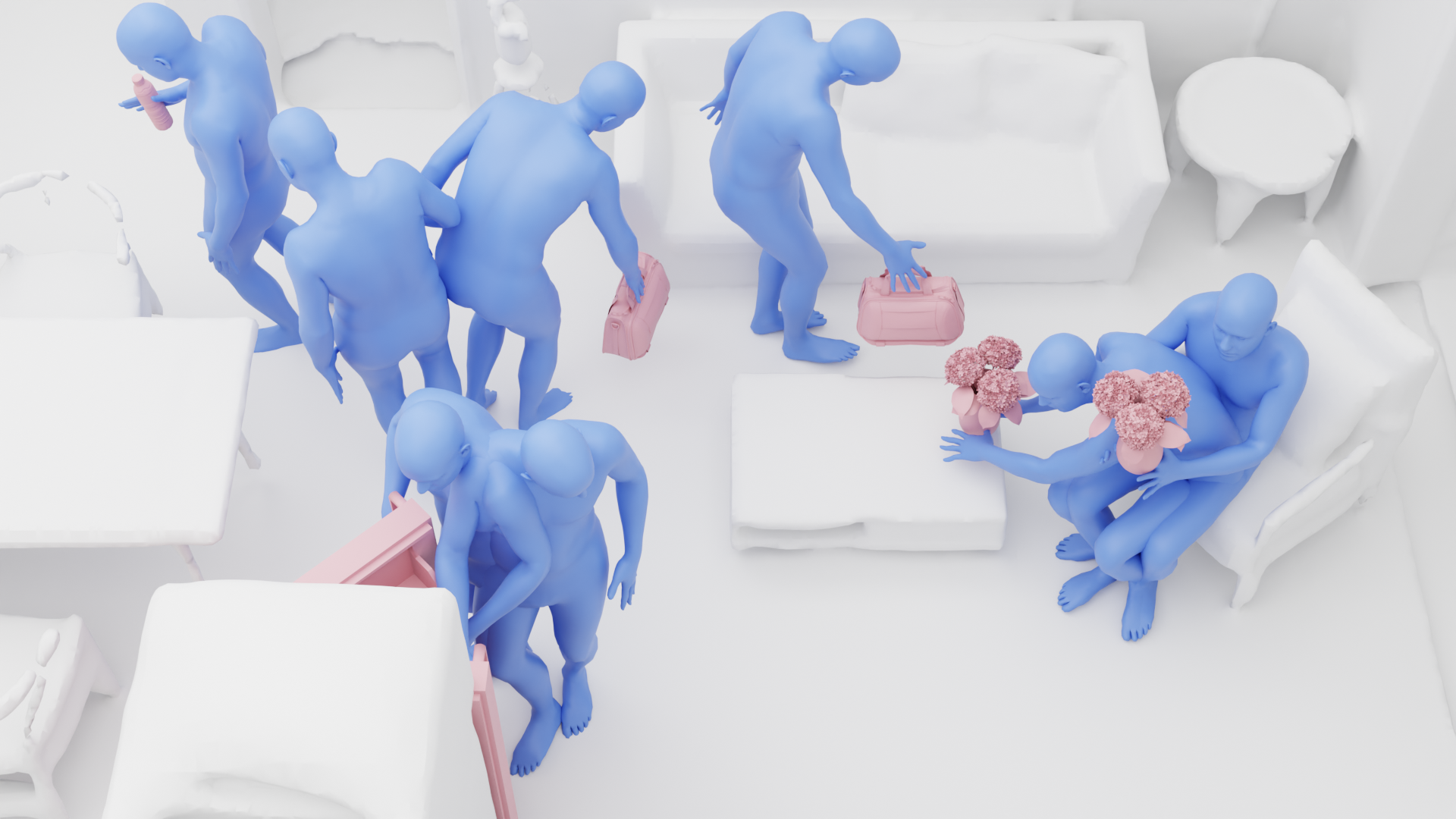} \\
      (a)&(b)
   \end{tabular}
   \caption{\textbf{Our Task.} Our method takes 3D object motion and a 3D scene as input (a), to synthesize physically consistent 3D human motion interacting with both the moving object and the static background scene (b).}
   \label{fig:task}
\end{figure*}
\section{Introduction}


Human motion synthesis is essential to the success of many applications like VR/AR, animation, and embodied AI, and has witnessed significant progress in recent years~\cite{lucas2022posegpt,petrovich2021action,chen2023executing,Guo_2022_CVPR,karunratanakul2023guided,petrovich2022temos,tevet2023human,zhang2023generating,guo2024momask,alexanderson2023listen,gong2023tm2d,li2021ai,siyao2022bailando,tseng2023edge, zeng2025chainhoi, xue2025guiding}. However, these methods overlook a key aspect of human activity, namely, interaction with the surroundings. In this paper, we address the problem of synthesizing human motion interacting with dynamic scenes i.e., scenes with a moving object, a more realistic setting for real applications, as illustrated in Fig.~\ref{fig:task}.

Although recent advances in generative models have inspired diffusion-based methods~\cite{xu2023interdiff,li2023object,li2023controllable,jiang2024scaling,xu2024interdreamer,lv2024himo, diller2024cg} to explore synthesizing human motion that interacts with dynamic objects, these methods often rely on simple interaction priors, such as contact or penetration constraints~\cite{xu2023interdiff, li2023controllable, xu2025interact, diller2024cg}, to encourage interaction. While effective in improving the quality of motion synthesis, such works do not accurately model the true physical forces, resulting in artifacts like floating, foot sliding, or unrealistic contact.

To enable physically valid human motion generation, physics simulators and reinforcement learning ~\cite{merel2020catch,hassan2023synthesizing,xie2023hierarchical, xu2025intermimic,wu2025human, wang2023physhoi} offer another direction by incorporating physical constraints to produce more realistic results. However, they rely on non-differentiable simulators which prevent seamless integration with end-to-end generative pipelines. To achieve physically plausible human motion in dynamic scenes within an end-to-end framework, PhysPT~\cite{zhang2024physpt} uses a continuous contact model for motion estimation. However, PhysPT~\cite{zhang2024physpt} is limited to motion estimation on static planar surfaces, assuming fixed surface normals and decoupled spring systems for normal and tangential forces, which cannot generalize to arbitrary geometries or capture the coupling between normal force and friction.

To overcome these limitations, we propose a novel continuous contact force model that can generalize to arbitrary, dynamic 3D surfaces and better reflects real-world contact physics. Specifically, 
our model decomposes the contact force into normal and tangential components. For the normal force, inspired by~\cite{brubaker2009estimating,zhang2024physpt}, we use a damped spring system. Instead of assuming a fixed upward-facing surface normal (opposite to the direction of gravity), we align it with the local surface geometry to handle arbitrary contact surfaces.~
We model the tangential force with separate static and kinetic components, with static friction proportional to lateral acceleration and kinetic friction proportional to the normal force.~This formulation ensures that normal and frictional forces are coupled and compatible with gradient-based optimization, enabling physically consistent human–object interactions in dynamic scenes. 

Given our defined continuous contact force models, we formulate human's dynamics with Euler–Lagrange equations. Specifically, we model the human's interaction not only with static scene but also dynamic objects. Crucially, unlike prior works~\cite{zhang2024physpt, li2023object, li2023controllable, xu2023interdiff}, we explicitly formulate the motion dynamics of the moving object with Euler–Lagrange equations. Following Newton’s third law, the force applied by the human to the object is equal and opposite to the reaction force applied by the object to the human. This allows us to integrate the object’s dynamics directly into that of the human, yielding a unified formulation where the human motion is constrained not only by the scene but also by the dynamics of the moving object.

To achieve physics-aware human motion synthesis, we thus propose a two-stage pipeline to integrate our proposed continuous contact model in an end-to-end learning framework for human motion synthesis. In the first stage, we propose to utilize a diffusion model to predict parameters for our contact force model and the human hand trajectory given the static scene and a moving object. They are further used as conditions to generate human motion with another diffusion model in the second stage. We introduce a loss defined by the Euler-Lagrange equations with our continuous contact model to encourage the physical consistency between the generated motion and the dynamic scene. Our contributions can be summarised as follows:
\begin{itemize}
    \item We introduce a novel continuous contact force model that accurately captures real contact forces and generalizes to arbitrary surfaces and geometries.
    \item We explicitly model dynamics of moving objects and integrate them into the human dynamics formulation based on Newton’s third law, ensuring that reciprocal forces between human and object are consistently preserved. 
    \item We propose a two-stage diffusion-based pipeline for scene-aware human motion synthesis that first predicts parameters for our new continuous contact model and then generates physics-aware human motion.
\end{itemize}
We evaluate our method on OMOMO dataset~\cite{li2023object} and TRUMANS dataset~\cite{jiang2024scaling} to demonstrate that our approach can achieve state-of-the-art performance and physically plausible results.

%% file: sec/03-related-work.tex
\section{Related Work}


\noindent{\bf Human-Scene and Human-Object Interaction.}  
The emergence of paired scene-aware~\cite{hassan2019resolving,li2023object, jiang2024scaling} and object-aware~\cite{li2023controllable, xu2025interact, huang2024intercap} datasets has fueled interest in synthesizing human motion that interacts realistically with scenes and objects. Existing methods~\cite{hassan2021stochastic,wang2021synthesizing,wang2022humanise,mao2022contact,xing2024scene} generate scene-aware human motion by leveraging geometric constraints such as collision avoidance, contact priors, and distance field.
However, they do not explicitly model interaction physics, often resulting in artifacts such as floating or foot skating.~RL-based approaches~\cite{merel2020catch,hassan2023synthesizing,xie2023hierarchical, pan2025tokenhsi} improved realism but are task-specific, hard to generalize, and non-differentiable.
Human-object interaction synthesis initially targeted small, handheld objects, but later expanded to full-body datasets~\cite{li2023object,xu2024interdreamer,jiang2024scaling, li2024genzi}.~Recent works~\cite{bhatnagar2022behave,xu2023interdiff,li2023object,li2023controllable,jiang2024scaling,jia2025primhoi} incorporated contact or penetration priors to enable dynamic interactions, yet they cannot predict forces, often leading to unrealistic artifacts.~RL-based policies~\cite{hassan2023synthesizing, xu2025intermimic} have been applied to larger objects but still lack generalization. While some works~\cite{tripathi20233d, zhang2024physpt, tripathi2024humos, shao2025finephys, zhang2024incorporating, gartner2022differentiable} incorporate differentiable physics into human modeling via body shape conditioning, biomechanical stability, or Lagrangian formulations, they primarily focus on body-level dynamics and interactions with a fixed ground plane, and do not explicitly model interaction forces with external objects or complex scenes.

In contrast, we propose a physics-aware human motion synthesis framework that jointly models full-body motion and physically plausible contact forces in cluttered 3D environments with dynamic objects. Our method integrates a physically grounded, differentiable continuous contact force model, generalizable to arbitrary surfaces, into a two-stage diffusion pipeline: predicting physics parameters and then generating motion. This formulation avoids hard contact assumptions and brittle RL policies, achieving realistic, physically consistent human–scene–object interactions in a differentiable, end-to-end learnable manner.

\noindent{\bf Contact Modelling.} 
Previous Human-object interaction synthesis methods often rely on binary contact labels, distance thresholds, or vertex correspondences to model contact~\cite{xie2025intertrack, chen2023detecting, tripathi2023deco, xie2023visibility, cseke2025pico}, focusing mainly on the hands ~\cite{xu2023interdiff, li2023object, li2023controllable}. While these approaches encourage plausible contact patterns, they fail to capture continuous interaction forces, leading to artifacts like floating or sliding. Early works~\cite{mordatch2013animating, mordatch2012contact} directly optimize contact variables or forces, which are not well-suited for learning-based generation. PhysPT ~\cite{zhang2024physpt} introduced a continuous contact model using damped springs, but it assumes planar surfaces and decouples normal and tangential forces, limiting its generalizability.

In contrast, we propose a differentiable continuous contact model that aligns with local surface geometry and explicitly couples normal and frictional forces. This enables physically consistent HOI across arbitrary 3D surfaces and dynamic environments, and integrates seamlessly into generative pipelines.

%% file: sec/04-method.tex
\section{Approach}
Let us now introduce our approach to human motion synthesis in a dynamic environment consisting a moving object and a static scene as background.~Following the setup in~\cite{li2023object,jiang2024scaling}, we assume a given object motion $\mathbf{O}\in\mathbb{R}^{T\times B}$ and a static scene represented by 3D voxels $\mathbf{S}\in\{0,1\}^{N_x\times N_y\times N_z}$, where $T$ is the number of frames, $B$ is the object translation and BPS representation following \cite{li2023object} and $(N_x,N_y,N_z)$ defines the volume resolution. Our goal is then to generate a human motion $\mathbf{Q}\in\mathbb{R}^{T\times D}$, where $D$ is the human motion dimension that interacts with the moving object and the scene. 
In this section, we first formalize the dynamics of human and object motion, then introduce our continuous contact force model, and finally describe our two-stage pipeline that leverages physical principles to generate realistic human motion.


\subsection{Preliminary}
We introduce preliminaries on human and object dynamics below. It details the mathematical formulation of human motion and human's interactions with the scene in the physical world via Euler-Lagrange equations and prepares the ground for our contact modeling.


\noindent\textbf{Human Motion Dynamics.} In the context of modeling human dynamics, a human body is often considered as an object with multiple rigid parts and modeled with rigid body dynamics. Following previous works~\cite{li2023object,zhang2024physpt}, we use the popular SMPL~\cite{loper2023smpl} human model. In SMPL model, a human is represented by a pose parameter $\boldsymbol{\theta}\in\mathbb{R}^{23\times3}$, a shape parameter $\boldsymbol{\beta}\in\mathbb{R}^{10}$, and a global orientation  $\mathbf{R}\in\mathbb{R}^3$, $\mathbf{T}\in\mathbb{R}^3$. Let's denote the human pose as 
\begin{equation}
    \mathbf{q} = \{\boldsymbol{\theta},\mathbf{R},\mathbf{T}\}\in\mathbb{R}^{75}\;.
    \label{eq:smpl_pose}
\end{equation}
The Euler-Lagrange Equations for the motion of such a human body are then defined as 
\begin{equation}
\mathbf{M}_h(\mathbf{q}) \ddot{\mathbf{q}} + \mathbf{C}_h(\mathbf{q}, \dot{\mathbf{q}}) + \mathbf{G}_h(\mathbf{q}) 
= \boldsymbol{\tau} + \mathbf{J}_{hs}^\top \boldsymbol{\lambda}_{s} + \mathbf{J}_{ho}^\top \boldsymbol{\lambda}_{o}\;,
\label{eq:human-dynamics}
\end{equation}
where $\dot{\mathbf{q}} \in \mathbb{R}^{75}$ and $\ddot{\mathbf{q}} \in \mathbb{R}^{75}$ are the velocities and accelerations of human joints, respectively. $\mathbf{M}_h(\mathbf{q}) \in \mathbb{R}^{75 \times 75}$ represents the human mass matrix that depends on the body mass and segmental inertias. $\mathbf{C}_h \in \mathbb{R}^{75}$ and $\mathbf{G}_h \in \mathbb{R}^{75}$ capture Coriolis/centrifugal and gravitational effects, respectively. On the right-hand side of the equation, $\boldsymbol{\tau} \in \mathbb{R}^{75}$ denotes internal joint torques from e.g., muscles. $\boldsymbol{\lambda}_s \in \mathbb{R}^{3C_s}$ denote the external contact forces from the static scene to human body, and $\boldsymbol{\lambda}_{o}\in\mathbb{R}^{3C_o}$ represent the contact force from the moving object to human hand. $C_s$ and $C_o$ are the number of contact points on human body and hand respectively. $\mathbf{J}_{hs} \in \mathbb{R}^{3C_s \times 75}$ and $\mathbf{J}_{ho}\in\mathbb{R}^{3C_o\times 75}$ are the contact Jacobian matrix that maps the contact forces to the forces on human joints. 

\noindent\textbf{Object Motion Dynamics.} Similarly, the motion dynamics of the moving object can be formulated as
\begin{equation}
\mathbf{M}_o(\mathbf{q}_o) \ddot{\mathbf{q}}_o + \mathbf{C}_o(\mathbf{q}_o, \dot{\mathbf{q}}_o) + \mathbf{G}_o(\mathbf{q}_o) 
= -\mathbf{J}_{o}^\top \boldsymbol{\lambda}_{o}\;,
\label{eq:object-dynamics}
\end{equation}
where $\mathbf{q}_o \in \mathbb{R}^{6}$ denotes the object orientation i.e., rotation and translation, and its velocity and acceleration are represented as $\dot{\mathbf{q}}_o \in \mathbb{R}^{6}$ and $\ddot{\mathbf{q}}_o \in \mathbb{R}^{6}$, respectively. $\mathbf{M}_o(\mathbf{q}_o)\in \mathbb{R}^{6 \times 6}$ is the object's mass matrix. $\mathbf{C}_o \in \mathbb{R}^{6}$ is the Coriolis and centrifugal forces, and $\mathbf{G}_o \in \mathbb{R}^{6}$ represents the gravitational forces. $\mathbf{J}_o \in \mathbb{R}^{3C_o \times 6}$ is the contact Jacobian matrix.  
Importantly, the contact force $\boldsymbol{\lambda}_{o}$ here appears with the opposite sign compared to Eq.~\ref{eq:human-dynamics}, reflecting Newton’s third law: the force that the human hand exerts on the object is equal in magnitude but opposite in direction to the force the object exerts on the hand. This reciprocal relationship tightly couples the object dynamics to the human dynamics. By leveraging this property, we can not only capture how the object responds to human manipulation, but also infer the corresponding reaction forces acting on the human hand. This coupling provides essential constraints for modeling physically consistent human–object interactions.

Note that, the left hand side in Eq.~\ref{eq:human-dynamics} and Eq.~\ref{eq:object-dynamics}, is only related to the subject motion, and its intrinsics such as mass, inertia. The contact Jacobian matrices on the right hand side of those equations can be precomputed from a set of possible contact points on the human body and the object. The problem becomes how to model the contact forces. In~\cite{brubaker2009estimating,zhang2024physpt}, a continuous contact force model is proposed to capture the contact force by two orthogonal spring systems as shown in Fig.~\ref{fig:contact}(a). Those two independent spring systems do not account for the coupling between the normal and tangential components of contact force and can only model the contact force between the human feet and the ground. To address this, we propose to explicitly model the tangential component as the static friction force and the kinetic friction force, while for the normal component, we still use the spring system. Furthermore, our formulation allows to compute the contact force on arbitrary surfaces. In the next section, we will introduce such contact force model.  

\subsection{Continuous Contact Force Model}

\begin{figure}[!htbp]
    \centering
    \begin{tabular}{cc}
          \includegraphics[width=0.4\linewidth]{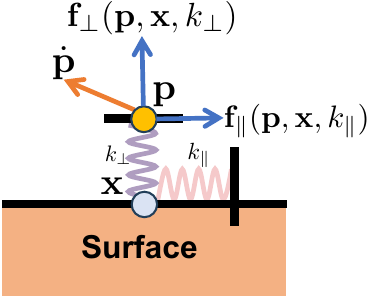}& \includegraphics[width=0.4\linewidth]{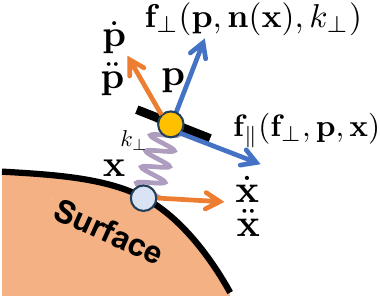}\\
          (a)  & (b) 
    \end{tabular}
    \vspace{-0.2cm}
    \caption{\textbf{Continous contact force model.} a) The PhysPT model assumes a static ground plane and represents contact force with two independent orthogonal springs, 
    b) Our model generalizes to arbitrary 3D surfaces by incorporating local surface normals for the normal force and explicitly modeling tangential static and kinetic friction that are dependent to the normal force, enabling physically consistent interactions in dynamic scenes.}
    \vspace{-0.5cm}
    \label{fig:contact}
\end{figure}

To capture human dynamics, it is essential to model both joint actuations and contact forces.~Modeling contact forces is particularly challenging because the contact status is often unknown and difficult to estimate accurately. Moreover, discrete contact representations introduce non-differentiable processes in force estimation. Although PhysPT~\cite{zhang2024physpt} mitigates this issue by adopting a continuous contact model inspired by a spring-mass system, it relies on unrealistic assumptions, such as infinite planar surfaces and globally upward-facing normals, and non-physical force directions in the horizontal plane, which limits its applicability. To overcome these limitations, we propose a physically grounded  continuous contact force model. 

As shown in Fig.~\ref{fig:contact}(b), given a possible contact point on human body $\mathbf{p}$, and its nearest surface point $\mathbf{x}$ on the moving object or the static scene, let's denote their relative position as $\Tilde{\mathbf{p}} = \mathbf{p} - \mathbf{x}$. We further decompose such relative position into normal and tangential components as,
\[
    \Tilde{\mathbf{p}}_{\perp} = (\Tilde{\mathbf{p}}^\top\mathbf{n}(\mathbf{x}))\mathbf{n}(\mathbf{x})\;,\;\; \;
    \Tilde{\mathbf{p}}_{\|} = \Tilde{\mathbf{p}} -  \Tilde{\mathbf{p}}_{\perp}\;,
\]
where $\mathbf{n}(\mathbf{x})$ is the surface normal at $\mathbf{x}$. The contact force between human and the object/scene is then defined as\footnote{Here, the contact forces are exerted on the human body by the static scene or moving object.}
\begin{equation}
    \boldsymbol{\lambda}(\mathbf{p}) = h(-\alpha (\|\Tilde{\mathbf{p}}\|-d_0)) h(\beta (\Tilde{\mathbf{p}}^\top\mathbf{n}(\mathbf{x}) + d_1))\mathbf{f}(\mathbf{p})\;,
    \label{eq:contact_force}
\end{equation}
where \( h(x) = \frac{1}{1 + e^{-x}} \) is a soft gating function. $\alpha>0,\;\beta>0$ are hyperparameters that regulate the transition sharpness. The other two hyperparameters $d_0$ and $d_1$ control the contact buffer. $
\mathbf{f}(\mathbf{p})$ denotes the force from the object/scene to the human. Under such definition, our contact forces will only be activated when human body is close enough to the surface while does not severely penetrates the surface i.e., $\|\Tilde{\mathbf{p}}\|$ is small, and $\Tilde{\mathbf{p}}^\top\mathbf{n}(\mathbf{x})$ is positive or small negative.

We further decompose the force to normal and tangential components as $\mathbf{f}_{\perp}$ and $\mathbf{f}_{\|}$, respectively,
\begin{equation}
     \mathbf{f}(\mathbf{p}) = \mathbf{f}_{\perp}(\mathbf{p}) + \mathbf{f}_{\|}(\mathbf{p})\;.
\end{equation}
For the normal component of the contact force, we follow~\cite{zhang2024physpt,brubaker2009estimating} to model it with a damped spring system. Unlike their formulation, which can only model forces from the ground where the normal vector always points upward, our formulation incorporates surface normals, enabling us to model forces from arbitrary surfaces. Formally, it is defined as
\begin{equation}
    \mathbf{f}_{\perp}(\mathbf{p}) = k(\mathbf{p}) \mathbf{n}(\mathbf{x})\;,
\end{equation}
where $k(\mathbf{p})$ is defined as
\begin{equation}
    k(\mathbf{p}) = -\kappa(\|\Tilde{\mathbf{p}}_{\perp}\| - d_0) - \delta (\dot{\Tilde{\mathbf{p}}}^\top \mathbf{n}(\mathbf{x})),
    \label{eq:normal_force}
\end{equation}
,~$\kappa>0$ and $\delta>0$ are stiffness and damping coefficients of the spring, $\dot{\Tilde{\mathbf{p}}}=\dot{\mathbf{p}} - \dot{\mathbf{x}}$ defines relative velocity between human contact point and its nearest object/scene point. 

To capture the tangential component, previous methods~\cite{zhang2024physpt,brubaker2009estimating} also propose to use the damped spring system. However, such system ignores the fact that the tangential component i.e., the friction can depends on the normal force. To address this issue, we propose a drastically different strategy to directly model the static ($\mathbf{f}_s$) and kinetic ($\mathbf{f}_k$) friction.
\begin{equation}
    \mathbf{f}_{\|}(\mathbf{p}) = \mathbf{f}_s(\mathbf{p}) + \mathbf{f}_k(\mathbf{p})\;,
\end{equation}
where the static friction term $\mathbf{f}_s(\mathbf{p})$ is defined as:
\begin{equation}
    \mathbf{f}_s(\mathbf{p}) = h(-\gamma(\|\dot{\Tilde{\mathbf{p}}}\| - v_0)) \rho |\|\Tilde{\mathbf{p}}_{\|}\| - d_0|\mathbf{d}_{\|},
\end{equation}
$\mathbf{d}_{\|}\in\mathbb{R}^{3}$ is the tangential force direction. 

Due to the different force analysis of static scenes and moving objects, such tangential force direction is also computed differently. In particular, for a static scene, such direction is defined as the acceleration of contact point projecting onto the tangential plane,
\vspace{-0.5cm}
\begin{equation}
    \mathbf{d}_{\|} = \frac{\ddot{\mathbf{p}}_{\|}}
    {\|\ddot{\mathbf{p}}_{\|}\|}\;,
\end{equation}
where $\ddot{\mathbf{p}}_{\|}$ is the tangential acceleration of the human contact point. Such formulation reflects the fact that the static friction from a static scene supports the motion of the human. For example, during walking, the overall external force that drives the human to move forward is from the ground to the supporting foot. However, for a moving object, it is the opposite i.e., human moves the object. So that, in this scenario, the tangential direction is opposite to the acceleration caused by external forces acting on the moving object, excluding the gravity.
\vspace{-0.5cm}
\begin{equation}
   \mathbf{d}_{\|} = -\frac{(\mathbf{a} - \mathbf{g})_{\|}}
    {\|(\mathbf{a} - \mathbf{g})_{\|}\|}\;
\end{equation}
where $\mathbf{a}$ is the acceleration of the object. $\mathbf{g}$ is gravitational acceleration. $(\mathbf{a} - \mathbf{g})_{\|}=(\mathbf{a} - \mathbf{g}) - ((\mathbf{a} - \mathbf{g})^\top\mathbf{n}(\mathbf{x}))\mathbf{n}(\mathbf{x})$ is the projection of overall acceleration of the moving object onto the tangent plane.

\begin{figure*}[!htbp]
    \centering
    \includegraphics[width=0.80\linewidth]{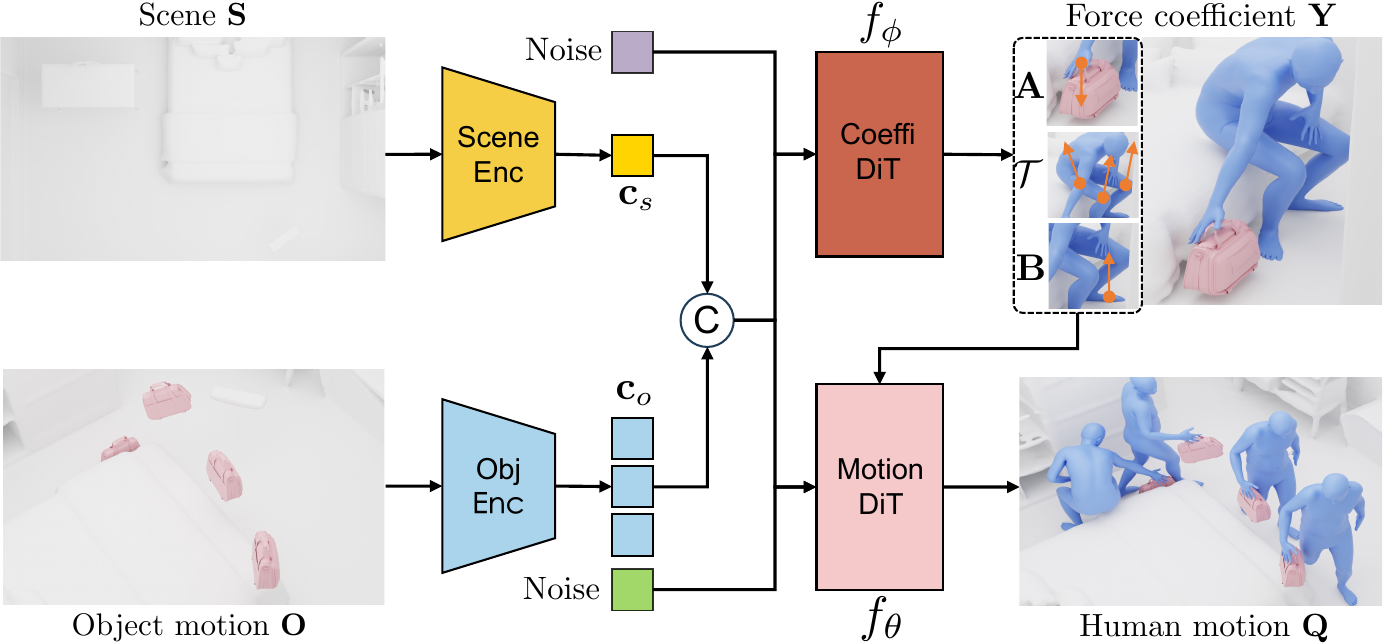}
    \vspace{-0.3cm}
\caption{
Overview of our pipeline. 
The input static scene $\mathbf{S}$ and object motion $\mathbf{O}$ is encode to a scene token $\mathbf{c}_s$ and several motion tokens $\mathbf{c}_o$.  In stage~1 (Top) a transformer-based diffusion model predicts force coefficients, namely, joint torques $\mathcal{T}$, contact parameters $\mathbf{A}, \mathbf{B}$ and hand trajectories $\mathbf{H}$.  In the next stage (Bottom), conditioned on such coefficients, the scene and object motion tokens, another transformer-based diffusion model is employed to generate the human motion that interacts with the scene and object. Thanks to our continuous contact modeling, we can employ a dynamic loss to further encourage physically valid interaction based on Eq.~\ref{eq:human-dynamics}. }
    \label{fig:pipeline}
    \vspace{-0.4cm}
\end{figure*}
Following the physics law, we define the kinetic friction to be linearly related to the normal force and its direction is opposite to the relative velocity, i.e.,
\begin{equation}
    \mathbf{f}_k(\mathbf{p}) = -\mu \|\mathbf{f}_{\perp}(\mathbf{p})\|
    \frac{ \dot{\Tilde{{\bf p}}}_{\|} }
    { \| \dot{\Tilde{{\bf p}}}_{\|}\|},
\end{equation}
where \( \mu>0 \) is the kinetic friction coefficient and $\dot{\Tilde{{\bf p}}}_{\|} $ is the projection of relative velocity onto the tangent plane.

In our continuous contact force model, the contact force is formulated as a function of motion, scene geometry, involving model coefficients ($\kappa,\delta,\rho,\mu$). Note that, since the scene contact varies across time steps and human contact points, those coefficients differ for each contact point at different frame. Let us denote coefficients for $C_s$ human body contact points and $C_o$ hand contact points in a $T$ frame sequence as $\mathbf{A}\in\mathbb{R}^{T\times C_s\times4}$ and $\mathbf{B}\in\mathbb{R}^{T\times C_o\times4}$, respectively.
We then design a two-stage pipeline whose first stage generates those model coefficients and the human joint torques for our continuous contact force model. The second stage takes model coefficients and joint torques as input to further synthesize human motion. A physics-aware loss is adopted to encourage the consistency between the coefficients and the human motion.

\subsection{Our Pipeline}
Having established the physical modeling of forces between humans, objects, and scenes, we now describe how these components are integrated into a unified motion synthesis framework. As shown in Fig.~\ref{fig:pipeline}, our pipeline consists of two stages namely, force coefficients generation and physics-aware human motion synthesis. 
Specifically, given the static scene $\mathbf{S}$ in voxel representation and object motion $\mathbf{O}$, we follow~\cite{jiang2024scaling,li2023object}, to first encode the scene and object motion to a $C$ dimensional scene token $\mathbf{c}_{s}\in\mathbb{R}^{C}$ and object motion tokens $\mathbf{c}_{o}\in\mathbb{R}^{T\times C}$, respectively.

\noindent{\bf  Force Coefficients Generation.}  
The first stage of our pipeline employs a diffusion-based network to generate the force coefficients from the scene and object motion tokens. The force coefficients include the internal joint torques $\mathcal{T}\in\mathbb{R}^{T\times 75}$, the force model coefficients for human contact points $\mathbf{A}$ and $\mathbf{B}$.  Additionally, we follow previous work~\cite{li2023object}, to also generate the hand trajectories $\mathbf{H} \in \mathbb{R}^{T \times 6}$, which is proved to be effective in producing accurate hand-object interaction.
More formally, we follow~\cite{li2023object} to design a transformer-based diffusion model as,
\begin{equation}
    \hat{\mathbf{Y}}_0 = f_{\phi}(\mathbf{Y}_n, \mathbf{c}_s,\mathbf{c}_o,n)\;,
\end{equation}
where $\hat{\mathbf{Y}}_0 = \{\hat{\mathbf{H}}, \hat{\mathcal{T}},\hat{\mathbf{A}}, \hat{\mathbf{B}}\}$ is the predicted force coefficients and the hand position.  $\mathbf{Y}_n$ is the noisy force coefficients, $n$ is the diffusion step. $f_\phi$ is the transformer diffusion model. The model is trained with $\ell_1$ loss,

\begin{equation}
\mathcal{L}_{\text{diff}} =
\mathbb{E}_{\mathbf{y}_0, n} \big[
\| \hat{\mathbf{Y}}_0 - \mathbf{Y}_0 \|_1
\big]\;,
\end{equation}
where $\mathbf{Y}_0=\{\mathbf{H},\mathcal{T},\mathbf{A},\mathbf{B}\}$ is the ground truth coefficients. 

\begin{table*}[t]
\centering
\vspace{-0.1cm}
\caption{Comparison of methods on the OMOMO dataset. Lower is better for error metrics; higher is better for precision, recall, and F1.}
\vspace{-0.3cm}
\begin{tabular}{lccccc|ccccc}
\toprule
Methods & HandJPE & MPJPE & MPVPE & $T_\text{root}$ & $O_\text{root}$ & Coll.(\%) & FS & $C_\text{prec}$ & $C_\text{rec}$ & F1 \\
\midrule
OMOMO~\cite{li2023object}                & 24.01 & 12.42 & 16.67 & 18.44 & 0.50 & 0.22 & 0.38 & 0.82 & 0.70 & 0.72 \\
Interdiff~\cite{xu2023interdiff}            & 31.76 & 16.03 & 20.19 & 21.50 & 0.61 & 0.20 & \textbf{0.35} & 0.69 & 0.58 & 0.63 \\
CHIOS~\cite{li2023controllable}                & 28.50 & 14.96 & 18.73 & 19.48 & 0.56 & 0.24 & 0.39 & 0.72 & 0.63 & 0.67 \\
InterAct~\cite{xu2025interact} &24.62 &12.59& 16.71 &18.8 & 0.47 & 0.21 & 0.39 &0.82 &0.71&0.73\\
\midrule
Ours                 & \textbf{20.09} & \textbf{10.02} & \textbf{13.60} & \textbf{16.07} & \textbf{0.40} & \textbf{0.19} & 0.41 & \textbf{0.87} & \textbf{0.76} & \textbf{0.80} \\
\bottomrule
\end{tabular}
\label{tab:omomo}
\vspace{-0.3cm}
\end{table*}


\begin{table*}[t]
\centering
\caption{Comparison of methods on the TRUMANS dataset. }

\begin{tabular}{lccccc|ccccc|c}
\toprule
Methods & HandJPE & MPJPE & MPVPE & $T_\text{root}$ & $O_\text{root}$ & Collision & FS & $C_\text{prec}$ & $C_\text{rec}$ & F1 & Sc. Pen. \\
\midrule
Trumans~\cite{jiang2024scaling}   & 47.85 & 36.20 & 38.02 & 33.90 & 0.45 & 0.25 & \textbf{0.39} & 0.65 & 0.54 & 0.59 & 33.48 \\
\midrule
Ours & \textbf{38.00} & \textbf{31.28} & \textbf{34.20} & \textbf{29.30} & \textbf{0.41} & \textbf{0.23} & 0.42 & \textbf{0.75} & \textbf{0.66} & \textbf{0.69} & \textbf{21.03} \\
\bottomrule
\end{tabular}
\label{tab:trumans}
\vspace{-0.3cm}
\end{table*}

\noindent{\bf Physics-aware Human Motion Synthesis.}  Given the force coefficients and hand position,  the second stage of our pipeline aims to generate the full-body human motion $\mathbf{Q}$. Similarly, we use another transformer-based diffusion model in this stage,

\begin{equation}
    \hat{\mathbf{Q}}_0 = f_{\theta}(\mathbf{Q}_n,\hat{\mathbf{Y}}_0, \mathbf{c}_s, \mathbf{c}_o,  n)\;, 
\end{equation}

where $\mathbf{Q}_n$ is the noisy human motion at $n$-th diffusion step. We train the model with the $\ell_1$ loss,
\begin{equation}
    \mathcal{L}_{\text{reco}} =  \mathbb{E}_{\mathbf{Q}_n,n} [\| \hat{\mathbf{Q}}_0 - \mathbf{Q}_0 \|_1]\;.
\end{equation}
where $\mathbf{Q}_0\in\mathbb{R}^{T\times 75}$ is ground truth human motion.
To further encourage the generated human motion to be physically valid, we define a dynamic consistency loss using Eq.~\ref{eq:human-dynamics} as
\begin{align}
    \mathcal{L}_{\text{dyn}} =
    \sum_{t=1}^{T} \| 
    & \mathbf{M}_h(\hat{\mathbf{q}}_t) \ddot{\hat{\mathbf{q}}}_t + \mathbf{C}_h(\hat{\mathbf{q}}_t, \dot{\hat{\mathbf{q}}}_t) + \mathbf{G}_h(t) \nonumber\\
    & - \mathbf{J}_{hs}(t)^\top \lambda_{s}(\mathbf{a}_t) - \mathbf{J}_{ho}(t)^\top \lambda_{o}(\mathbf{b}_t) - \boldsymbol{\tau}_t 
    \|_1,
    \nonumber
\end{align}
where $\hat{\mathbf{q}}_t, \dot{\hat{\mathbf{q}}}_t, \ddot{\hat{\mathbf{q}}}_t$ denote the predicted human pose, velocity and acceleration, respectively. Velocities and accelerations are computed from the predicted pose sequence.
 $\mathbf{a}_t\in\mathbb{R}^{C_s\times4}$ is the parameters for the contact between human body and scene. $\mathbf{b}_t\in\mathbb{R}^{C_o\times4}$ represents the parameters for the contact between human hand and object. 

Finally, the total loss for the second stage combines both objectives:
\begin{equation}
    \mathcal{L} =
    \mathcal{L}_{\text{reco}} + \lambda_{\text{dyn}} \mathcal{L}_{\text{dyn}}\;,
\end{equation}
where $\lambda_{\text{dyn}}>0$ is a balancing weight.

%% file: sec/05-experiments.tex
\section{Experiments}

\begin{figure*}
\centering
\setlength{\tabcolsep}{0pt} 
\renewcommand{\arraystretch}{0} 

\begin{tabular}{cccccc}
\scriptsize Ground Truth & \scriptsize Ours & \scriptsize OMOMO ~\cite{li2023object} & \scriptsize CHOIS ~\cite{li2023controllable} & \scriptsize InterDiff~\cite{xu2023interdiff} & \scriptsize InterAct~\cite{xu2025interact}  \\

\includegraphics[width=0.16\textwidth]{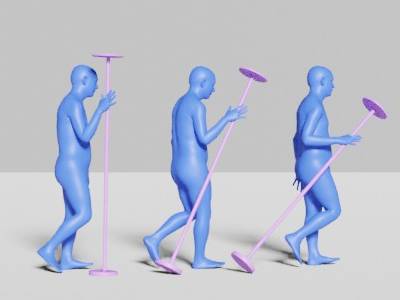} 
& \includegraphics[width=0.16\textwidth]{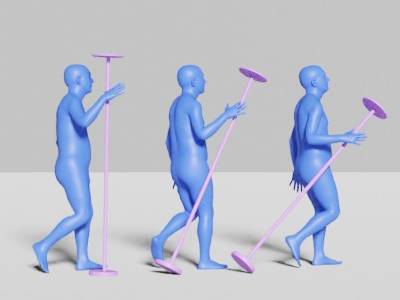} 
& \includegraphics[width=0.16\textwidth]{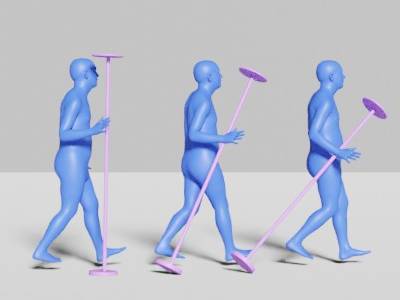} 
& \includegraphics[width=0.16\textwidth]{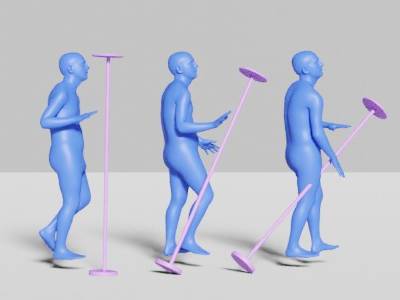} 
& \includegraphics[width=0.16\textwidth]{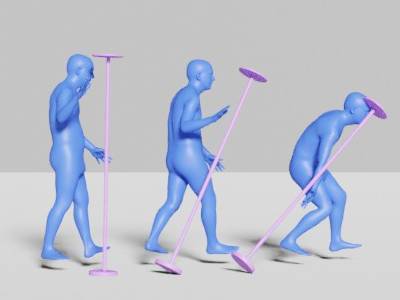}
& \includegraphics[width=0.16\textwidth]{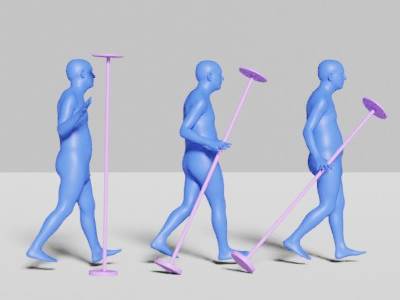}\\

 \includegraphics[width=0.16\textwidth]{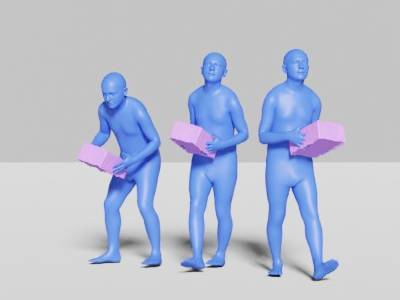} 
& \includegraphics[width=0.16\textwidth]{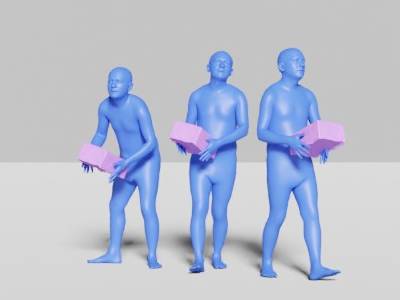} 
& \includegraphics[width=0.16\textwidth]{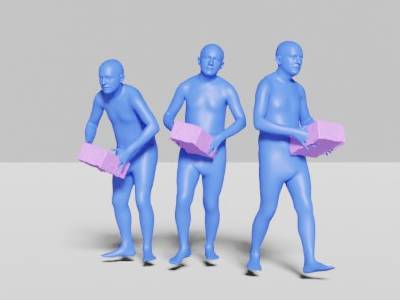} 
& \includegraphics[width=0.16\textwidth]{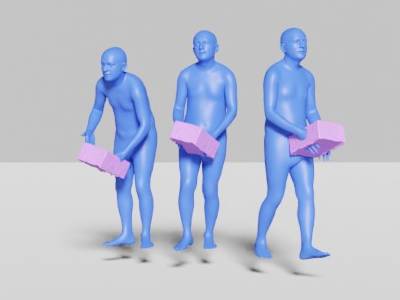} 
& \includegraphics[width=0.16\textwidth]{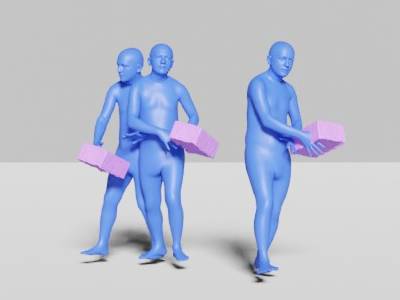}
& \includegraphics[width=0.16\textwidth]{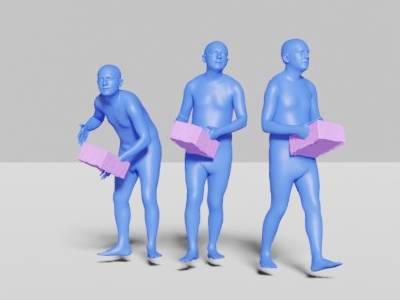}\\
\end{tabular}
\vspace{-0.1cm}
\caption{Qualitative comparison on OMOMO. From left to right: object-only context, ground truth, our prediction, and predictions from OMOMO, CHOIS, InterDiff, and InterAct.}
\label{fig:omomo}
\vspace{-0.3cm}
\end{figure*}

\begin{figure*}
\centering
\includegraphics[width=0.9\textwidth]{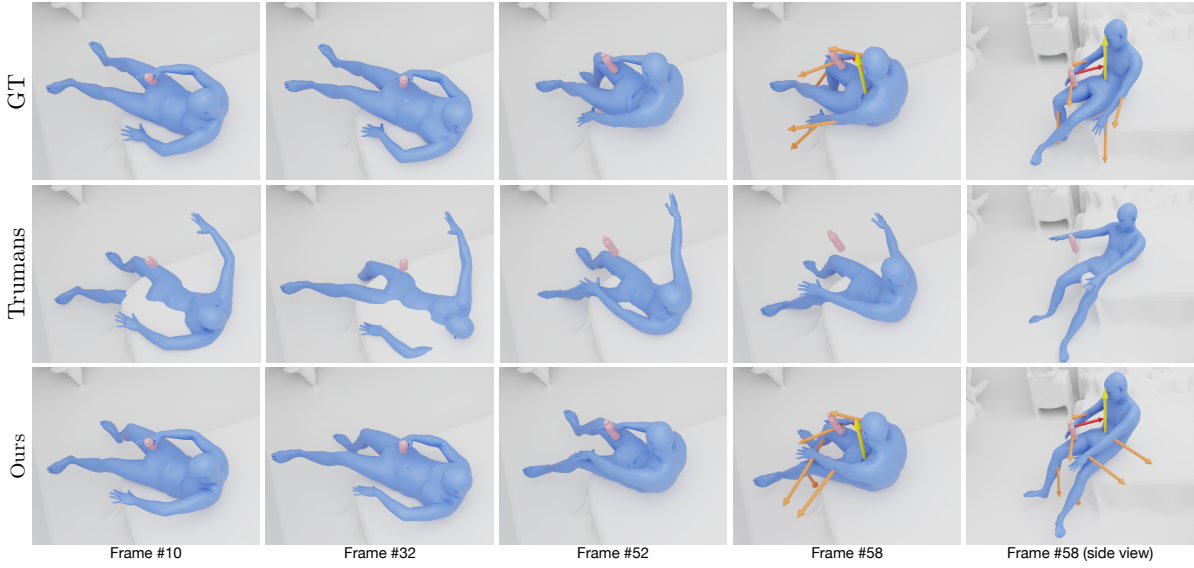}
\vspace{-0.3cm}
\caption{Qualitative comparison on Trumans. Each row shows ground truth, the Turmans baseline, and our method. Arrows illustrate estimated forces: \textcolor{red}{red} for forces from the dynamic object to the human, \textcolor{yellow}{yellow} for forces from the static scene to the human, and \textcolor{orange}{orange} for internal joint forces.}
\label{fig:trumans}
\vspace{-0.5cm}
\end{figure*}

\subsection{Datasets}
We evaluate our method on two recent human-scene interaction datasets: OMOMO dataset~\cite{li2023object} and TRUMANS dataset~\cite{jiang2024scaling}.

\noindent\textbf{OMOMO.}
The OMOMO dataset~\cite{li2023object} is a high-quality motion capture dataset featuring approximately 10 hours of paired human–object interactions on a fixed horizontal plane, involving 15 everyday objects. 
SMPL-X parameters were extracted via MoSh++, and object poses were estimated using markers attached to the objects. Our setting are following the protocol in~\cite{li2023object}.
In this dataset, our method accounts for both the interaction between the hands and the moving object, and the interaction between the feet and the ground plane.

\noindent\textbf{TRUMANS.} The TRUMANS datase~\cite{jiang2024scaling} is a large-scale corpus for human–scene interaction modeling, containing ~15 hours of motion data (about 1.6M frames at 30 Hz) in richly populated indoor environments, with 20 common object categories. 
Since we focus on human motion synthesis in a dynamic environment, we use a subset of the dataset to make sure each sequence contains one dynamic object.

\subsection{Data Preparation}
To train and evaluate our framework, we require ground-truth parameters of the continuous contact force model, such as joint torques and contact coefficients. Given the ground-truth human and object motion sequences, we obtain the internal joint torques $\mathcal{T}\in\mathbb{R}^{T\times75}$ and the contact model coefficients $\mathbf{A}, \mathbf{B}$ via dynamic optimization following~\cite{zhang2024physpt}. Specifically, we leverage the human and object dynamics formulations in Eq.~\ref{eq:human-dynamics} and Eq.~\ref{eq:object-dynamics} to minimize the residuals of the Euler–Lagrange equations.
This optimization assumes known constants such as gravity $\mathbf{g}$, SMPL-derived segment masses and inertias (regressed from $\boldsymbol{\beta}$), object mass and inertia, and a predefined set of candidate contact points on both the human body and objects. Under these assumptions, we solve for $\mathbf{A}, \mathbf{B}, \mathcal{T}$ to ensure physically consistent supervision signals. Details of the optimization procedure, including implementation and solver settings, are provided in the supplementary material.

\subsection{Evaluation Metrics}


\noindent\textbf{Metrics for human motion.}
We adopt the evaluation protocol of~\cite{li2023object} to measure the quality of motion using HandJPE, MPJPE, MPVPE, root translation error ($T_\text{root}$), root orientation error ($O_\text{root}$), and foot sliding (FS). HandJPE, MPJPE, and MPVPE correspond to the mean hand joint position error, mean per-joint position error, and mean per-vertex position error (in cm), respectively. The root translation error is computed as the Euclidean distance between predicted and ground-truth root positions, while the root orientation error is defined as the Frobenius norm between the predicted and ground-truth rotation matrices, i.e., $\| \mathbf{R}_\text{pred} \mathbf{R}_\text{gt}^{-1} - I \|_2$. We compute foot sliding (FS) following~\cite{he2022nemf}.

\noindent\textbf{Metrics for human object interaction.}
We evaluate human object interactions via Collision Percentage and contact metrics ($C_\text{prec}$, $C_\text{rec}$, F1) same as~\cite{li2023object}. Collision Percentage is the percentage of frames where the synthesized human mesh penetrates the object. The threshold for penetration is 4cm. 
Contact metrics follow the object detection protocol: we use a 5 cm threshold between hands and the object mesh to obtain the binary contact labels for both the predicted human meshes and the ground truth ones. We then report the precision ($C_\text{prec}$), recall ($C_\text{rec}$), and F1 score for the contact labels.

\noindent\textbf{Metric for human scene interaction.}
Following~\cite{jiang2024scaling}, we also report the scene penetration metric to measure the interaction between the human and the static scene.

\subsection{Baselines}
For the OMOMO dataset, we compare our method with four recent baselines on human motion synthesis. \textbf{OMOMO}~\cite{li2023object} synthesizes human motion conditioned on object trajectories.
\textbf{InterDiff}~\cite{xu2023interdiff} generates human–object interactions given motion histories.
\textbf{CHOIS}~\cite{li2023controllable} generates human and object motions from language descriptions, initial states, and waypoints. \textbf{InterAct}~\cite{xu2025interact} generates human motion based on object sequences.
For fair comparison, we adapt InterDiff~\cite{xu2023interdiff} and CHOIS~\cite{li2023controllable} to generate human motion conditioned solely on object motion, and InterAct~\cite{xu2025interact} using SMPL-X as representation. For OMOMO~\cite{li2023object}, we directly include the results reported in their paper for comparison.

For the TRUMANS dataset, we use~\cite{jiang2024scaling} as our baseline. While~\cite{jiang2024scaling} generates human and object motion from scene, text, action labels, and goals, we adapt it to take the 3D scene and object motion as input for human motion synthesis, using the same object motion encoding as our method.

\begin{figure*}[t]
\centering
\setlength{\tabcolsep}{0pt} 
\renewcommand{\arraystretch}{0} 
\begin{tabular}{ccccccc}
\scriptsize Object & \scriptsize Ground Truth & \scriptsize Ours & \scriptsize Ours w/o PL &  \scriptsize Ours w/o OBJ &  \scriptsize Ours w PT \\
\includegraphics[width=0.16\textwidth]{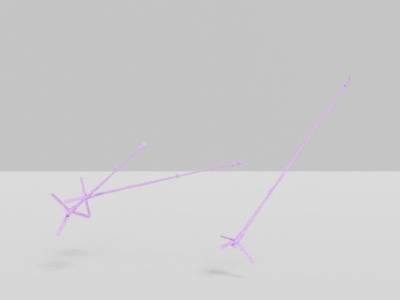} 
& \includegraphics[width=0.16\textwidth]{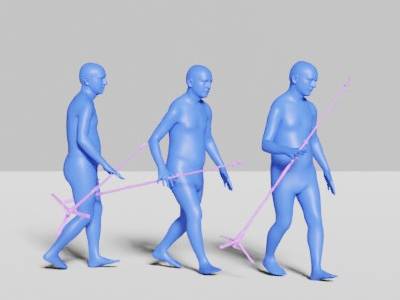} 
& \includegraphics[width=0.16\textwidth]{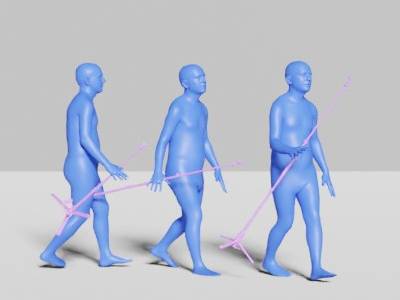} 
& \includegraphics[width=0.16\textwidth]{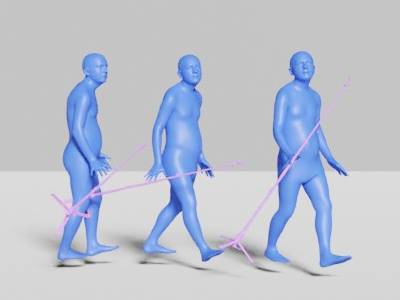} 
& \includegraphics[width=0.16\textwidth]{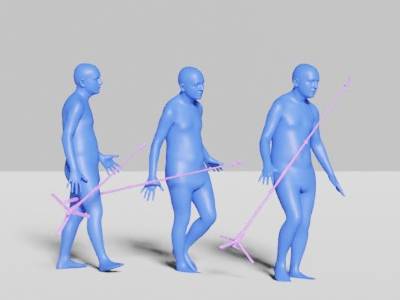}  
& \includegraphics[width=0.16\textwidth]{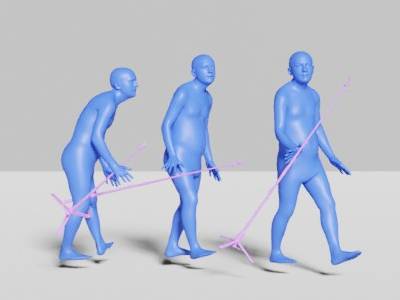} \\
\end{tabular}
\vspace{-0.3cm}
\caption{Ablation study comparison on OMOMO. }
\label{fig:ablation}
\vspace{-0.3cm}
\end{figure*}
\begin{table*}[t]
\centering
\caption{Ablation study of our method. Lower is better for error metrics; higher is better for precision, recall, and F1.}
\vspace{-0.3cm}
\begin{tabular}{lccccc|ccccc}
\toprule
Methods & HandJPE & MPJPE & MPVPE & $T_\text{root}$ & $O_\text{root}$ & Collision & FS & $C_\text{prec}$ & $C_\text{rec}$ & F1 \\
\midrule
Ours w/o PL & 21.63 & 11.18 & 14.05 & 18.33 & 0.45 & 0.21 & 0.43 & 0.81 & 0.67 & 0.73 \\
Ours w/o OBJ & 21.02 & 10.73 & 13.89 & 16.52 & 0.43 & 0.22 & 0.43 & 0.80 & 0.69 & 0.74\\
Ours w PT        & 21.47 & 10.74 & 13.96 & 16.83 & 0.43 & 0.21 & 0.44 & 0.85 & 0.70 & 0.76 \\
Ours                 & \textbf{20.09} & \textbf{10.02} & \textbf{13.60} & \textbf{16.07} & \textbf{0.40} & \textbf{0.19} & \textbf{0.41} & \textbf{0.87} & \textbf{0.76} & \textbf{0.80} \\
\bottomrule
\end{tabular}
\label{tab:ablation}
\vspace{-0.5cm}
\end{table*}

\subsection{Implementation Details}
All models are implemented in PyTorch~\cite{paszke2019pytorch} and trained using the Adam optimizer~\cite{kingma2014adam} with an initial learning rate of 0.002. All experiments are conducted on a single NVIDIA RTX 4090 GPU.

\subsection{Results}
\noindent\textbf{OMOMO.} In Tab.~\ref{tab:omomo}, we compare our results to those of baselines. Our method consistently outperforms baseline methods for all metrics except for the foot sliding (FS).  Although our method has higher foot sliding score, as also discussed in~\cite{li2023object}, due to the definition of foot sliding, a lower foot sliding score does not necessarily mean a better results. One can achieve a very low foot sliding score to generate a human motion that floating above the ground. The qualitative comparison shown in Fig.~\ref{fig:omomo} also evidences this. For human motion metrics, our method achieves at least 12\% better performance across all evaluation metrics comparing to the second best model, i.e., OMOMO~\cite{li2023object}. For human object interaction, our results penetrate less with the object while have more precise and complete contact with it. Due to the lack of explicit physics modeling, the baseline methods always produces human motions with feet floating above the ground while the human motion from our method shows a better contact relationship between the feet and the ground. More qualitative results are shown in the supplementary video. 

\noindent\textbf{TRUMANS.} We report the results on the TRUMANS dataset~\cite{jiang2024scaling} in Tab.~\ref{tab:trumans}. The conclusion remains the same. Our method consistently outperforms the baseline method on all metrics. In this dataset, the static scenes have various geometries. The human scene interaction is not only between human feet and the ground but also between other human body parts and the scene surfaces, e.g., between the bottom of the human and the chair. We also show the qualitative comparison in Fig.~\ref{fig:trumans}. The human motion generated by our method have more realistic interaction with the scene and the object. More qualitative results are shown in the supplementary video.

\subsection{Ablation Study}
We conduct ablation studies on the OMOMO dataset~\cite{li2023object} to analyze the contributions of our components. Specifically, we evaluate:
(i) our method without the dynamic consistency loss (“Ours w/o PL”), our method without the object dynamic (“Ours w/o OBJ”) and (iii) our method using the contact model from PhysPT~\cite{zhang2024physpt} (“Ours w/ PT”).
The results are summarized in Tab.~\ref{tab:ablation}. Without dynamic consistency loss, the model performs consistently worse especially for the object contact, having a decease of 7\% on F1 score. Although with the contact model proposed in PhysPT~\cite{zhang2024physpt}, the overall performance becomes better than that of “Ours w/o PL”, it still consistently underperforms our method.


%% file: sec/06-conclusion.tex
\vspace{-0.3cm}
\section{Conclusion}
In this paper, we introduced a continuous contact model for dynamic scene aware human motion synthesis. Our formulation overcomes key limitations of existing contact models—most of which operate only on the ground plane and fail to capture realistic friction dynamics—by incorporating surface-normal-conditioned force modeling in the normal direction and explicit static and kinetic friction modeling on the tangent plane. We integrated this contact model into a two-stage, diffusion-based human motion synthesis pipeline that first predicts physics parameters and subsequently generates human motion conditioned on those physics parameters. Our approach achieves state-of-the-art performance on human motion generation in a dynamic scene, demonstrating the importance of explicit physics reasoning for high-fidelity interaction synthesis. Looking forward, we aim to extend our framework to more general and complex scenarios, including interactions involving multiple dynamic objects, articulated tools, and even multi-human collaborative behaviors under diverse environmental contexts. Such extensions would further expand the applicability of physics-aware motion synthesis in embodied AI.

\noindent{\bf Acknowledgements.} This research was supported partially by an ARC Discovery Grant (DP200102274).